\documentclass[runningheads]{llncs}
\usepackage{graphicx}

\usepackage{tikz}
\usepackage{comment}
\usepackage{amsmath,amssymb} %
\usepackage{color}
\usepackage{hyperref}
\usepackage[numbers,sort&compress]{natbib}
\usepackage{upgreek}

\usepackage{tabularx}
\newcolumntype{Y}{>{\centering\arraybackslash}X}

\usepackage[accsupp]{axessibility}  %

\usepackage{bm}
\newcommand{\myvec}[1]{\boldsymbol{\mathbf{#1}}}

\usepackage{layouts}

\usepackage{ulem}

\newcommand{\eg}{e.g.,~}
\newcommand{\etal}{et al.~}
\newcommand{\ie}{i.e.,~}

\definecolor{gold}{rgb}{0.83, 0.69, 0.22}
\definecolor{silver}{rgb}{0.79, 0.75, 0.73}
\definecolor{bronze}{rgb}{0.8, 0.5, 0.2}

\begin{document}
\pagestyle{headings}
\mainmatter
\def\ECCVSubNumber{6228}  %

\title{3D Face Reconstruction with Dense Landmarks}

\begin{comment}
\titlerunning{ECCV-22 submission ID \ECCVSubNumber} 
\authorrunning{ECCV-22 submission ID \ECCVSubNumber} 
\author{Anonymous ECCV submission}
\institute{Paper ID \ECCVSubNumber}
\end{comment}

\titlerunning{3D Face Reconstruction with Dense Landmarks}
\author{
\begin{tabularx}{\textwidth}{cYcY}
    Erroll Wood &
    Tadas Baltru\v{s}aitis &
    Charlie Hewitt &
    Matthew Johnson \\
    Jingjing Shen &
    Nikola Milosavljevi\'{c} &
    Daniel Wilde & 
    Stephan Garbin \\
    Chirag Raman &
    Jamie Shotton &
    Toby Sharp &
    Ivan Stojiljkovi\'{c} \\
    &
    Tom Cashman &
    Julien Valentin
    &
\end{tabularx}%
}
\authorrunning{E. Wood et al.}
\institute{Microsoft}
\maketitle

\begin{abstract}
Landmarks often play a key role in face analysis, but many aspects of identity or expression cannot be represented by sparse landmarks alone.
Thus, in order to reconstruct faces more accurately, landmarks are often combined with additional signals like depth images or %
techniques like differentiable rendering.
Can we keep things simple by just using more landmarks?
In answer, we present the first method that accurately predicts 10$\times$ as many landmarks as usual, covering the whole head, including the eyes and teeth.
This is accomplished using synthetic training data, which guarantees perfect landmark annotations.
By fitting a morphable model to these dense landmarks, we achieve state-of-the-art results for monocular 3D face reconstruction in the wild.
We show that dense landmarks are an ideal signal for integrating face shape information across frames by demonstrating accurate and expressive facial performance capture in both monocular and multi-view scenarios.
Finally, our method is highly efficient: we can predict dense landmarks and fit our 3D face model at over 150FPS on a single CPU thread.
Please see our website: \url{https://microsoft.github.io/DenseLandmarks/}.
\end{abstract}

\section{Introduction}

Landmarks are points in correspondence across all faces, like the tip of the nose or the corner of the eye.
They often play a role in face-related computer vision, %
\eg being used to extract facial regions of interest~\cite{hassner2014effective}, or helping to constrain 3D model fitting~\cite{Zollhofer2018review,garrido2018}.
Unfortunately, many aspects of facial identity or expression cannot be encoded by a typical sparse set of 68 landmarks alone.
For example, without landmarks on the cheeks, we cannot tell whether or not someone has high cheek-bones.
Likewise, without landmarks around the outer eye region, we cannot tell if someone is softly closing their eyes, or scrunching up their face.

\begin{figure}
    \includegraphics[width=\linewidth]{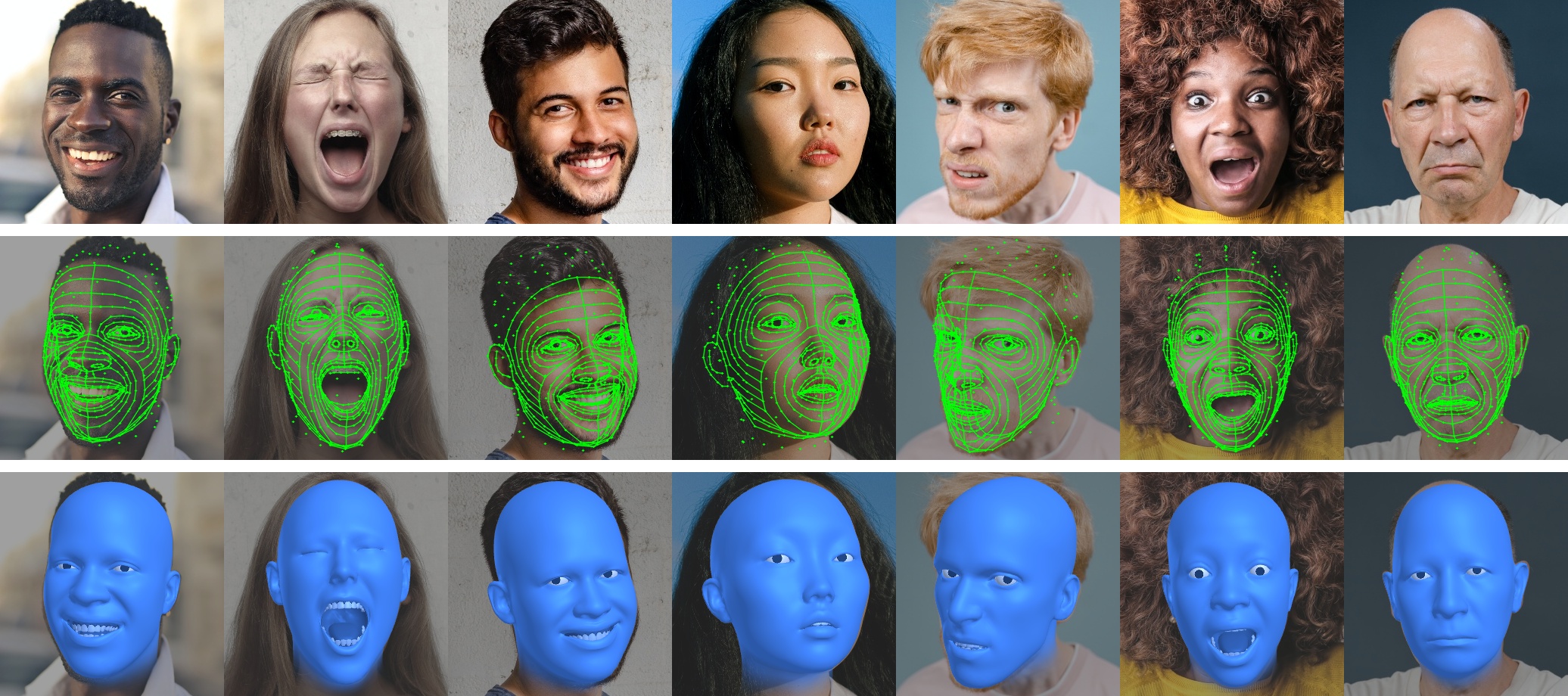}
    \caption{Given a single image (top), we first robustly and accurately predict 703 landmarks (middle). To aid visualization, we draw lines between landmarks. We then fit our 3D morphable face model to these landmarks to reconstruct faces in 3D (bottom).}
    \label{fig:intro-fig-1}
\end{figure}

In order to reconstruct faces more accurately, previous work has therefore used additional signals beyond color images, such as depth images~\cite{thies2015expression} or optical flow \cite{cao2018stabilized}.
However, these signals may not be available or reliable to compute.
Instead, given color images alone, others have approached the problem using analysis-by-synthesis: minimizing a photometric error~\cite{garrido2018} between a generative 3D face model and an observed image using differentiable rendering~\cite{dib2021diff,genova2018unsupervised}.
Unfortunately, these approaches are limited by the approximations that must be made in order for differentiable rendering to be computationally feasible.
In reality, faces are not purely Lambertian~\cite{feng2021learning}, and many important illumination effects are not explained using spherical harmonics alone~\cite{dib2021diff}, \eg ambient occlusion or shadows cast by the nose.

Faced with this complexity, wouldn't it be great if we could just use more landmarks?
We present the first method that predicts over 700 landmarks both accurately and robustly.
Instead of only the frontal ``hockey-mask'' portion of the face, our landmarks cover the entire head, including the ears, eyeballs, and teeth.
As shown in \autoref{fig:intro-fig-1}, these landmarks provide a rich signal for both facial identity and expression.
Even with as few as 68, it is hard for humans to precisely annotate landmarks that are not aligned with a salient image feature. %
That is why we use synthetic training data which guarantees consistent annotations.
Furthermore, instead of representing each landmark as just a 2D coordinate,
we predict each one as a random variable: a 2D circular Gaussian with position and uncertainty \cite{kendall2017uncertainties}.
This allows our predictor to express uncertainty about certain landmarks, \eg occluded landmarks on the back of the head.

Since our dense landmarks represent points of correspondence across all faces, we can perform 3D face reconstruction by fitting a morphable face model~\cite{blanz1999morphable} to them.
Although previous approaches have fit models to landmarks in a similar way~\cite{Zhu_2016_CVPR}, we are the first to show that landmarks are the only signal
required to achieve state-of-the-art results for monocular face reconstruction in the wild.

The probabilistic nature of our predictions also makes them ideal for fitting a 3D model over a temporal sequence, or across multiple views.
An optimizer can discount uncertain landmarks and %
rely on more certain ones. %
We demonstrate this with accurate and expressive results for both multi-view and monocular facial performance capture.
Finally, we show that predicting dense landmarks and then fitting a model can be highly efficient by demonstrating real-time facial performance capture at over 150FPS on a single CPU thread.

In summary, our main contribution is to show that you can achieve more with less.
You don't need parametric appearance models, illumination models, or differentiable rendering for accurate 3D face reconstruction.
All you need is a sufficiently large quantity of accurate 2D landmarks and a 3D model to fit to them.
In addition,
we show that combining probabilistic landmarks and model fitting lets us intelligently aggregate face shape information across multiple images by
demonstrating robust and expressive results for both multi-view and monocular facial performance capture.

\begin{figure}[t]
    \hfill \raisebox{1.5ex}{a)} \includegraphics[width=0.2\linewidth]{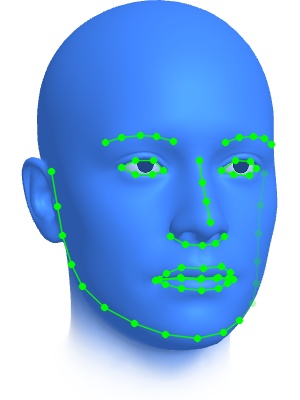}%
    \hfill \raisebox{1.5ex}{b)} \includegraphics[width=0.2\linewidth]{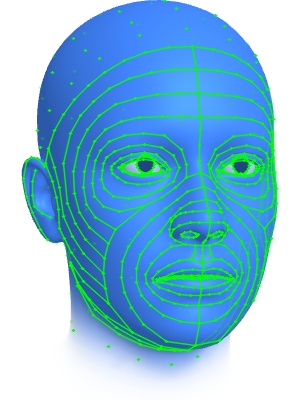}%
    \hspace{4pt}%
    \parbox[b][0.29\linewidth][c]{0.14\linewidth}{
        \includegraphics[width=\linewidth]{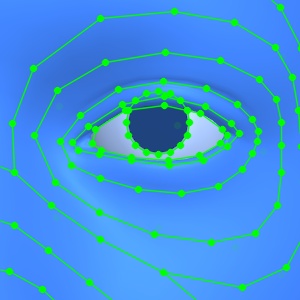}
    }%
    \hspace{7pt}%
    \parbox[b][0.29\linewidth][c]{0.14\linewidth}{
        \includegraphics[width=\linewidth]{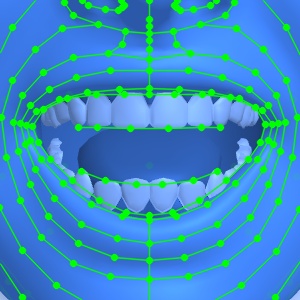}
    }%
    \hspace*{\fill}%
    \vspace{-1.5ex}
    \caption{Compared to a typical sparse set of 68 facial landmarks (a), our dense landmarks (b) cover the entire head in great detail, including ears, eyes, and teeth. These dense landmarks are better at encoding facial identity and subtle expressions.}
    \label{fig:method}
\end{figure}

\section{Related work}

Reconstructing faces in 3D from images is a mature field at the intersection of vision and graphics.
We focus our literature review on methods that are closer to our own, and
refer the reader to Morales \etal \cite{morales2021survey} for an extensive survey.

\subsubsection{Regression-based 3D face reconstruction}
DNN-based regression has been extensively used as a tool for 3D face reconstruction.
Techniques fall into two broad categories: supervised, and self-supervised.
Approaches either use 3D Morphable Models (3DMMs) \cite{blanz2003face, FLAME:SiggraphAsia2017,gerig2018morphable}, or eschew linear models and
instead learn a non-linear one as part of the training process \cite{tran2019towards}. 

Fully supervised techniques either use parameter values from a 3DMM
that is fit to the data via optimization as labels
\cite{chandran20semantic,tuan2017regressing,yi2019mmface},
or known face geometry is posed by sampling from a 3DMM and rendered to
create synthetic datasets \cite{dou2017end,richardson20163d,sela2017unrestricted,genova2018unsupervised}.
Self-supervised approaches commonly use landmark reprojection error and/or perceptual loss via differentiable rendering 
\cite{RingNet:CVPR:2019,deng2019accurate,feng2021learning,guo2020towards,Liu_2017_ICCV,guo2018cnn,genova2018unsupervised,richardson2017learning,tewari2018self,tewari2017mofa,tran2019towards,tran2018nonlinear}.
Other techniques augment this with 3D or multiview constraints
\cite{shang2020self,dou2018multi,liu2018disentangling,tewari2019fml,yoon2019self,zhou2019dense}.
While this is similar to our technique, we only use a DNN to regress landmark positions
which are then used to optimize 3DMM parameters, as in the large body of hybrid model-fitting methods \cite{Bogo:ECCV:2016,han2020megatrack}.

\subsubsection{Optimization-based 3D face reconstruction}
Traditionally, markerless reconstruction of face geometry is achieved with multi-view stereo \cite{SeitzMVSComparison, singleShotFacialCapture}, followed by optical flow based alignment, and then optimisation using geometric and temporal priors \cite{globallyConsistentReconstructionPopa, passiveFacialPerfCapture, anchorFramesPaper}.
While such methods produce detailed results, each step takes hours to complete.
They also suffer from drift and other issues due to their reliance on optical flow and multi-view stereo \cite{localGeometricFaceIndexingCong2019}.
While our method cannot reconstruct faces in such fine detail, it accurately recovers the low-frequency shape of the face, and aligns it with a common topology.
This enriches the raw data with semantics, making it useful for other tasks.

If only a single image is available, dense photometric \cite{dib2021diff,thies2016face2face}, depth  \cite{thies2015expression}, or optical flow \cite{cao2018stabilized} constraints are commonly used to recover face shape and motion.
However, these methods still rely on sparse landmarks for initializing the optimization close to the dense constraint's basin of convergence, and coping with fast head motion \cite{Zollhofer2018review}.
In contrast, we argue that dense landmarks alone are sufficient for accurately recovering the overall shape of the face.

\subsubsection{Dense landmark prediction}

While sparse landmark prediction is a mainstay of the field \cite{bulat2017far},
few methods directly predict dense landmarks or correspondences.
This is because annotating a face with dense landmarks %
is a highly ambiguous task, so either  synthetic data \cite{wood2021fake}, pseudo-labels made with model-fitting \cite{zhu2016face, Feng_2018_ECCV, deng2020retinaface}, or semi-automatic refinement of training data \cite{kartynnik2019real,jeni2015dense} are used.
Another issue with predicting dense landmarks is that
heatmaps, the \textit{de facto} technique for predicting landmarks \cite{bulat2017far,bulat2021subpixel}, rise in computational complexity with the number of landmarks.
While a few previous methods have predicted dense frontal-face landmarks via cascade regression \cite{jeni2015dense} or direct regression \cite{kartynnik2019real,grishchenko2020attention,deng2020retinaface}, we are the first to accurately and robustly predict over 700 landmarks covering the whole head, including eyes and teeth.

Some methods choose to predict dense correspondences as an image instead, where each pixel corresponds to a fixed point in a UV-unwrapping of the face \cite{alp2017densereg, Feng_2018_ECCV} or body \cite{guler2018densepose,taylor2012vitruvian}.
Such parameterization suffers from several drawbacks.
How does one handle self-occluded portions of the face, \eg the back of the head?
Furthermore, what occurs at UV-island boundaries? If a pixel is half-nose and half-cheek, to which does it correspond?
Instead, we choose to discretize the face into dense landmarks.
This lets us predict parts of the face that are self-occluded, or lie outside image bounds. %
Having a fixed set of correspondences also benefits the model-fitter, making it more amenable to running in real-time. %

\begin{figure}[t]
    \includegraphics[width=\linewidth]{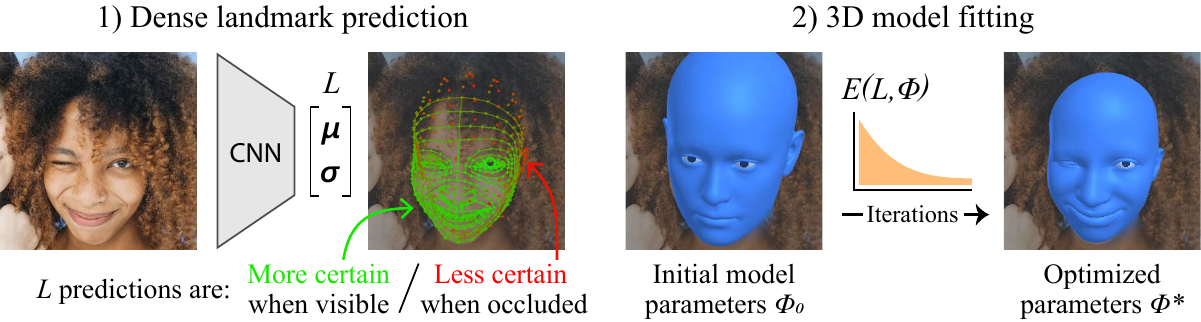}%
    \vspace{-1ex}%
    \caption{Given an image, we first predict probabilistic dense landmarks $L$, each with position $\mu$ and certainty $\sigma$.
    Then, we fit our 3D face model to $L$, minimizing an energy $E$ by optimizing model parameters $\myvec{\Phi}$.}
    \label{fig:method-1}
\end{figure}

\section{Method}

In recent years, methods for 3D face reconstruction have become more and more complicated, involving differentiable rendering and complex neural network training strategies. 
We show instead that success can be found by keeping things simple.
Our approach consists of two stages:
First we predict probabilistic dense 2D landmarks $L$ using a traditional convolutional neural network (CNN).
Then, we fit a 3D face model, parameterized by $\myvec{\Phi}$, to the 2D landmarks by minimizing an energy function $E(\myvec{\Phi}; L)$.
Images themselves are not part of this optimization; the only data used are 2D landmarks.

The main difference between our work and previous approaches is the number and quality of landmarks.
No one before has predicted so many 2D landmarks, so accurately.
This lets us achieve accurate 3D face reconstruction results by fitting a 3D model to these landmarks alone.

\subsection{Landmark prediction}

\subsubsection{Synthetic training data.}

\begin{figure}[t]
\includegraphics[width=\linewidth]{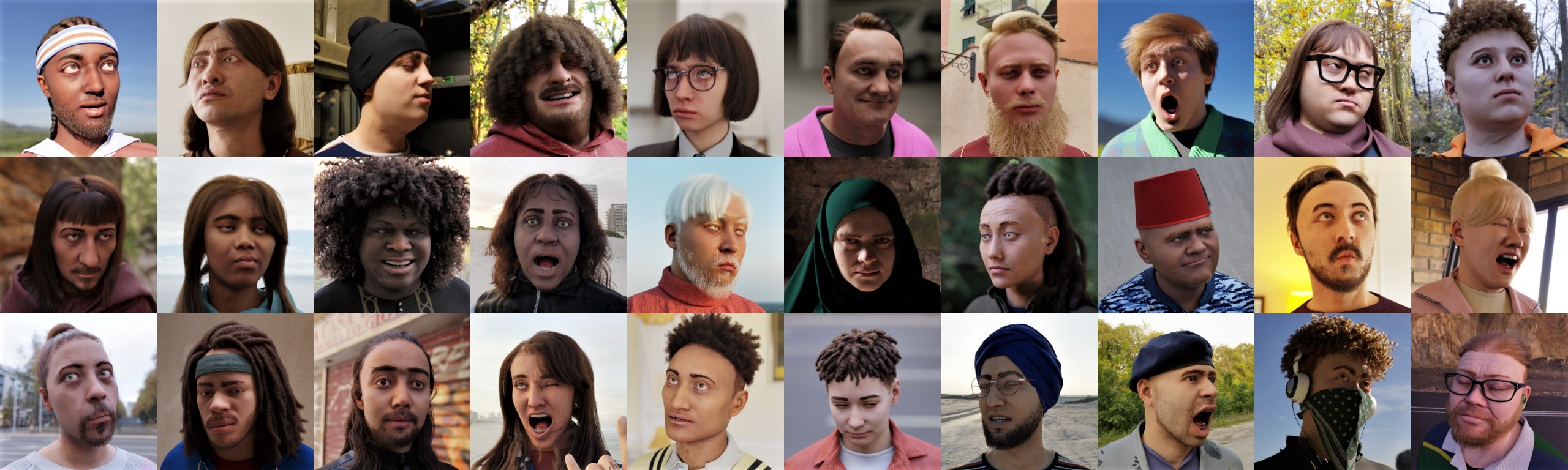}
\caption{Examples of our synthetic training data. Without the perfectly consistent annotations provided by synthetic data, dense landmark prediction would not be possible.}
\label{fig:synthetic_data}
\end{figure}

Our results are only possible because we use synthetic training data.
While a human can consistently label face images with \eg 68 landmarks,
it would be almost impossible for them to annotate an image with dense landmarks.
How would it be possible to consistently annotate occluded landmarks on the back of the head,
or multiple landmarks over a largely featureless patch of skin \eg the forehead?
In previous work, pseudo-labelled real images with dense correspondences are obtained by fitting a 3DMM to images \cite{alp2017densereg}, but the resulting label consistency heavily depends on the quality of the 3D fitting.
Using synthetic data has the advantage of guaranteeing perfectly consistent labels.
We rendered a training dataset of 100k images using the method of Wood et al. \cite{wood2021fake} with some minor modifications:
we include expression-dependent wrinkle texture maps for more realistic skin appearance,
and additional clothing, accessory, and hair assets.
See \autoref{fig:synthetic_data} for some examples.

\subsubsection{Probabilistic landmark regression.}

\begin{figure}[t]
    \includegraphics[width=0.326\linewidth]{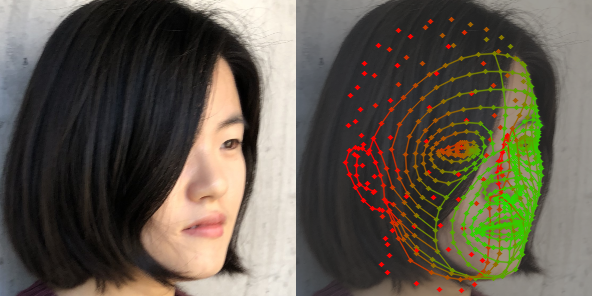} \hfill
    \includegraphics[width=0.326\linewidth]{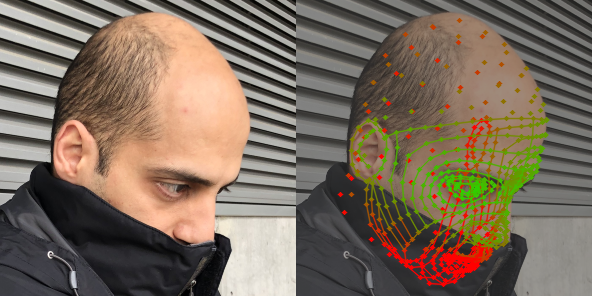} \hfill
    \includegraphics[width=0.326\linewidth]{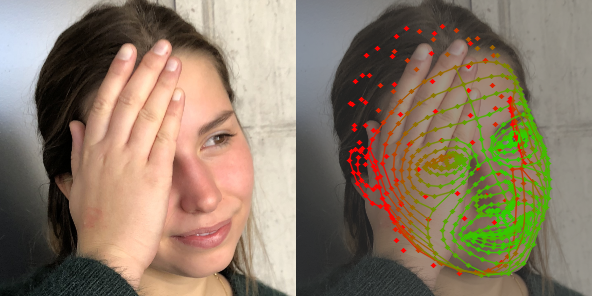}
    \caption{When parts of the face are occluded by e.g. hair or clothing, the corresponding landmarks are predicted with high uncertainty (red), compared to those visible (green).}
    \label{fig:method-occlusion-examples}
\end{figure}

We predict each landmark as a random variable with the probability density function of a circular 2D Gaussian.
So $L_i = \{\myvec{\mu}_i, \sigma_i\}$, where $\myvec{\mu}_i = [x_i, y_i]$ is the expected position of that landmark, and $\sigma_i$ (the standard deviation) is a measure of uncertainty.
Our training data includes labels for landmark positions $\myvec{\mu}_i^\prime = [x_i^\prime, y_i^\prime]$, but not for $\sigma$.
The network learns to output $\sigma$ in an unsupervised fashion to show that it is certain about some landmarks, \eg visible landmarks on the front of the face, and uncertain about others, \eg landmarks hidden behind hair (see \autoref{fig:method-occlusion-examples}).
This is achieved by training the network with a Gaussian negative log likelihood (GNLL) loss \cite{kendall2017uncertainties}:%
\begin{equation}
\textrm{Loss}(L) = \sum_{i=1}^{|L|} \lambda_i \Bigg(%
\underbrace{
    \vphantom{\frac{(x_i - x_i^\prime)^2 + (y_i - y_i^\prime)^2}{2 \sigma_i^2}} %
    \log\left(\sigma_i^2\right)}_{\textrm{Loss}_{\sigma}
} +
\underbrace{
    \frac{\| \myvec{\mu}_{i} - \myvec{\mu}_{i}^\prime \| ^2}{2 \sigma_{i}^2}
}_{\textrm{Loss}_{\mu}
}
\Bigg)
\label{eqn:loss}
\end{equation}
$\textrm{Loss}_{\sigma}$ penalizes the network for being too uncertain, and $\textrm{Loss}_{\mu}$ penalizes the network for being inaccurate.
$\lambda_i$ is a per-landmark weight that focuses the loss on certain parts of the face. %
This is the only loss used during training.

The probabilistic nature of our landmark predictions is important for accuracy.
A network trained with the GNLL loss is more accurate than a network trained with L2 loss on positions only.
Perhaps this is the result of the CNN being able to discount challenging landmarks (\eg fully occluded ones), and spend more capacity on making precise predictions about visible landmarks.

Landmarks are commonly predicted via heatmaps \cite{bulat2021subpixel}.
However, generating heatmaps is computationally expensive \cite{li20212d}; it would not be feasible to output over 700 heatmaps in real-time.
Heatmaps also prevent us predicting landmarks outside image bounds.
Instead, we keep things simple, and directly regress position and uncertainty using a traditional CNN.
We are able to take any off-the-shelf architecture, %
and alter the final fully-connected layer to output three values per-landmark: two for position and one for uncertainty.
Since this final layer represents a small percentage of total CNN compute, our method scales well with landmark quantity.

\textbf{Training details.}
Landmark coordinates are normalized from $[0, S]$ to $[-1, 1]$, for a square image of size $S \! \times \! S$.
Rather than directly outputting $\sigma$, we predict $\log \sigma$, and take its exponential to ensure $\sigma$ is positive.
Using PyTorch \cite{NEURIPS2019_9015}, we train ResNet \cite{he2016deep} and MobileNet V2 \cite{sandler2018mobilenetv2} models from the timm \cite{rw2019timm} library using AdamW \cite{ADAMW} with automatically determined learning rate \cite{falcon2019pytorch}.
We use data augmentation to help our synthetic data cross the domain gap \cite{wood2021fake}. %

\subsection{3D model fitting}

Given probabilistic dense 2D landmarks $L$, our goal is to find optimal model parameters $\myvec{\Phi}^*$ that minimize the following energy:
$$
E(\myvec{\Phi}; L) =
\underbrace{
    E_{
        \vphantom{identity}%
        \textrm{landmarks}%
    }
}_{
    \textrm{Data term}
}
+
\underbrace{
    E_{\textrm{identity}} +
    E_{\textrm{expression}} +
    E_{\textrm{joints}} +
    E_{\textrm{temporal}} +
    E_{\textrm{intersect}}
}_{
    \textrm{Regularizers}
}
$$
$E_{\textrm{landmarks}}$ is the only term that encourages the 3D model to explain the observed 2D landmarks.
The other terms use prior knowledge to regularize the fit.

Part of the beauty of our approach is how naturally it scales to multiple images and cameras.
In this section we present the general form of our method, suitable for $F$ frames over $C$ cameras, \ie multi-view performance capture.
\vspace{0.5em}

\textbf{3D face model.}
We use the face model described in \cite{wood2021fake}, comprising $N\!=\!7,\!667$ vertices and $K\!=\!4$ skeletal joints (the head, neck, and two eyes). %
Vertex positions are determined by the mesh-generating function
$\mathcal{M}(\myvec{\beta}, \myvec{\psi}, \myvec{\theta})
\!:\!\mathbb{R}^{|\myvec{\beta}|+|\myvec{\psi}|+|\myvec{\theta}|}\!\to\!\mathbb{R}^{3N}$
which takes parameters
$\myvec{\beta}\in\mathbb{R}^{|\myvec{\beta}|}$ for identity,
$\myvec{\psi}\in\mathbb{R}^{|\myvec{\psi}|}$ for expression, and
$\myvec{\theta}\in\mathbb{R}^{3K + 3}$ for skeletal pose (including root joint translation).
$$
\mathcal{M}(\myvec{\beta}, \myvec{\psi}, \myvec{\theta}) = 
\mathcal{L}(\mathcal{T}(\myvec{\beta}, \myvec{\psi}), \myvec{\theta}, \mathcal{J}(\myvec{\beta}); \mathbf{W})
$$
where $\mathcal{L}(\mathbf{V}, \myvec{\theta}, \mathbf{J}; \mathbf{W})$ is a standard linear blend skinning (LBS) function~\cite{Lewis2000skinning} that rotates vertex positions $\mathbf{V}\in\mathbb{R}^{3N}$ about joint locations $\mathbf{J}\in\mathbb{R}^{3K}$ by local joint rotations in $\myvec{\theta}$, with per-vertex weights $\mathbf{W}\in\mathbb{R}^{K\times N}$.
The face mesh and joint locations in the bind pose are determined by
$\mathcal{T}(\myvec{\beta}, \myvec{\psi})\!:\!\mathbb{R}^{|\myvec{\beta}| + |\myvec{\psi}|}\to\mathbb{R}^{3N}$
and 
$\mathcal{J}(\myvec{\beta})\!:\!\mathbb{R}^{|\myvec{\beta}|}\to\mathbb{R}^{3K}$
respectively.
See Wood et al.~\cite{wood2021fake} for more details.
\vspace{0.5em}

\begin{figure}[t]
\hfill%
a)%
\includegraphics[width=0.45\textwidth]{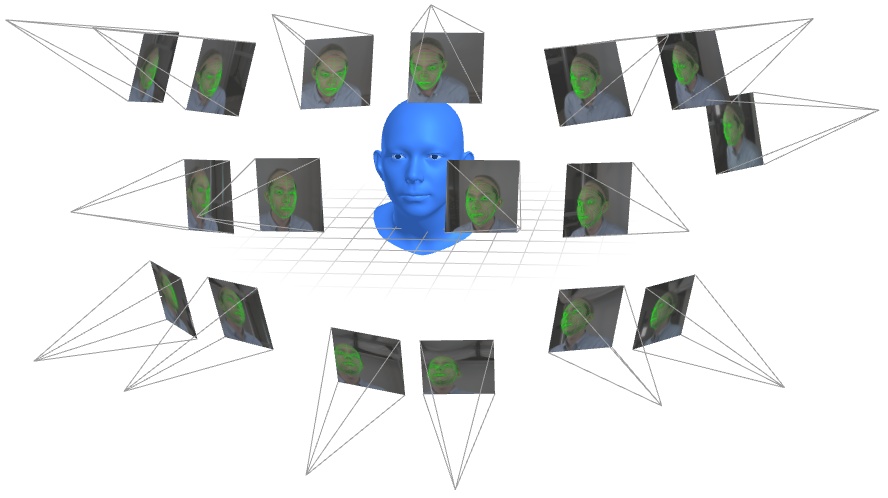}%
\hspace*{0.5em}%
b)
\includegraphics[width=0.4\textwidth]{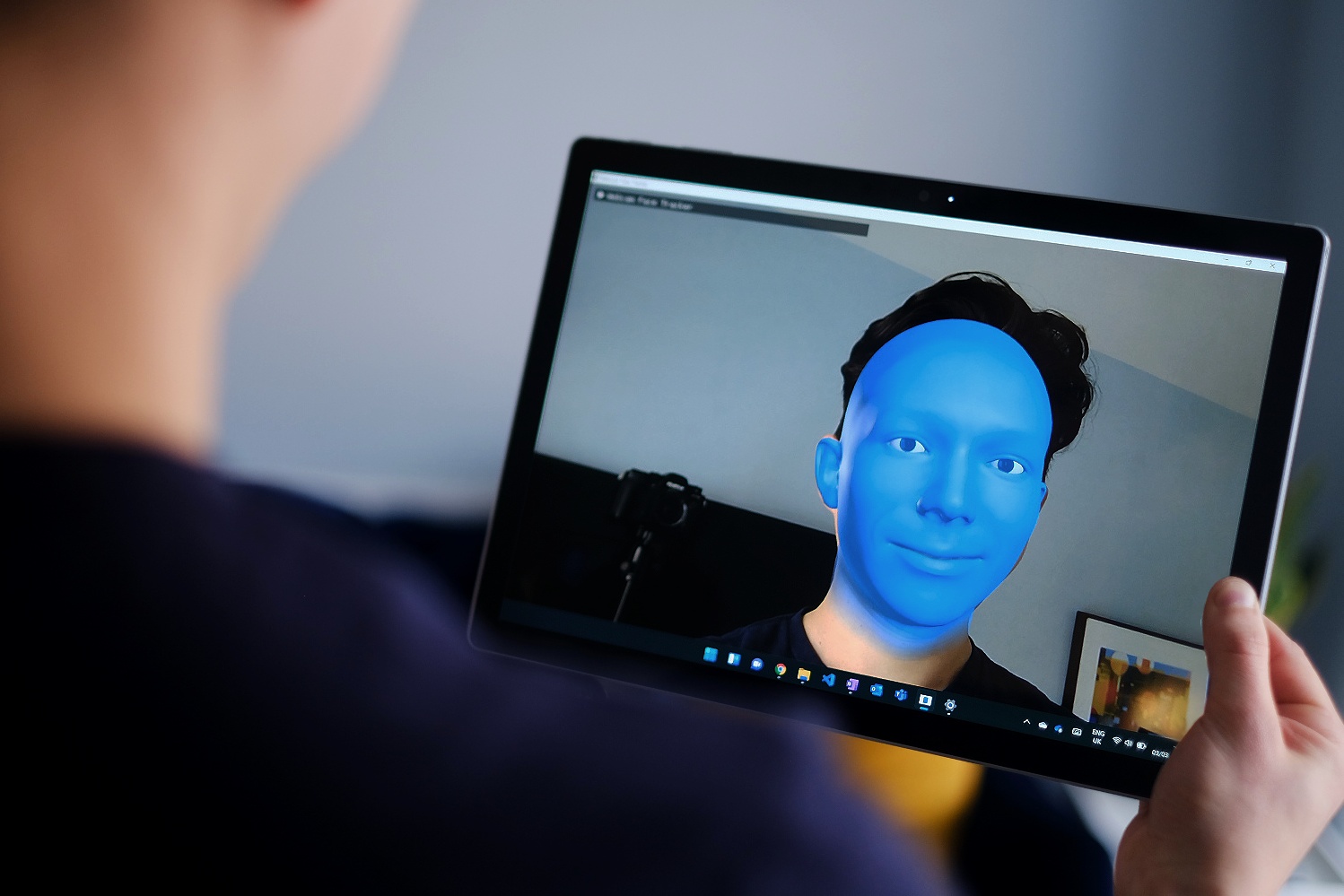}%
\hspace*{\fill}%
\caption{
We implemented two versions of our approach:
one for processing multi-view recordings offline (a), and one for real-time facial performance capture (b).}
\label{fig:implementations}
\end{figure}
\textbf{Cameras} are described by a world-to-camera rigid transform $\myvec{X} \in\mathbb{R}^{3 \times 4} = \left[ \myvec{R} | \myvec{T} \right]$ comprising rotation and translation, and a pinhole camera projection matrix $\myvec{\Pi} \in\mathbb{R}^{3 \times 3}$.
Thus, the image-space projection of the $j$\textsuperscript{th} landmark in the $i$\textsuperscript{th} camera is
$
\myvec{x}_{i,j} = \myvec{\Pi}_i \myvec{X}_i \mathcal{M}_j
$.
In the monocular case, $\myvec{X}$ can be ignored. %
\vspace{0.5em}

\textbf{Parameters} $\myvec{\Phi}$ are optimized to minimize $E$.
The main parameters of interest control the face, but we also optimize camera parameters if they are unknown.
\begin{equation*}
\myvec{\Phi} = \{
\underbrace{
    \myvec{\beta},
    \myvec{\Psi}_{F \times |\myvec{\psi}|},
    \myvec{\Theta}_{F \times|\myvec{\theta}|}
}_{\textrm{Face}}
;\,
\underbrace{
    \myvec{R}_{C \times 3},
    \myvec{T}_{C \times 3},
    \myvec{f}_{C}
}_{\textrm{Camera(s)}}
\}
\end{equation*}
Facial identity $\myvec{\beta}$ is shared over a sequence of $F$ frames, but expression $\myvec{\Psi}$ and pose $\myvec{\Theta}$ vary per frame.
For each of our $C$ cameras we have six degrees of freedom for rotation $\myvec{R}$ and translation $\myvec{T}$, and a single focal length parameter $f$.
In the monocular case, we only optimize focal length.

\textbf{$E_{\textrm{landmarks}}$} encourages the 3D model to explain the predicted 2D landmarks:
\begin{equation}
E_{\textrm{landmarks}} =
\sum_{i, j, k}^{F, C, |L|}
\frac{\| \myvec{x}_{ijk} - \myvec{\mu}_{ijk} \| ^2}{2 \sigma_{ijk}^2}
\label{eqn:e_landmarks}
\end{equation}
where, for the 
$k$\textsuperscript{th} landmark
seen by the $j$\textsuperscript{th} camera
in the $i$\textsuperscript{th} frame,
$[\myvec{\mu}_{ijk}, \sigma_{ijk}]$ is the 2D location and uncertainty predicted by our dense landmark CNN, and 
$\myvec{x}_{ijk} = \myvec{\Pi}_j \myvec{X}_j \mathcal{M}(\myvec{\beta}, \myvec{\psi}_i, \myvec{\theta}_i)_k$
is the 2D projection of that landmark on our 3D model.
The similarity of \autoref{eqn:e_landmarks} to $\textrm{Loss}_{\mu}$ in \autoref{eqn:loss} is no accident:
treating landmarks as 2D random variables during both prediction and model-fitting allows our approach to elegantly handle uncertainty,
taking advantage of landmarks the CNN is confident in,
and discounting those it is uncertain about.

\textbf{$E_{\textrm{identity}}$} penalizes unlikely face shape by maximizing the relative log-likeli\-hood of shape parameters $\myvec{\beta}$ under a multivariate Gaussian Mixture Model (GMM) of $G$ components
fit to a library of 3D head scans \cite{wood2021fake}.
$E_{\textrm{identity}} = -\log\left(p(\myvec{\beta})\right)$ where 
$
p(\myvec{\beta}) = \sum_{i=1}^{G}
\gamma_i\;
\mathcal{N}\!
\left(\myvec{\beta} | \myvec{\nu}_i,\, \myvec{\Sigma}_i\right)
$.
$\myvec{\nu}_i$ and $\myvec{\Sigma}_i$ are the mean and covariance matrix of the $i$\textsuperscript{th} component, and $\gamma_i$ is the weight of that component.

\textbf{$E_{\textrm{expression}}$} $= \| \myvec{\psi} \|^2$
and
\textbf{$E_{\textrm{joints}}$} $= \| \myvec{\theta}_{i:i\in[2,K]} \|^2$
encourage the optimizer to explain the data with as little expression and joint rotation as possible. We do not penalize global translation or rotation by ignoring the root joint $\myvec{\theta}_1$.
\vspace{0.5em}

\textbf{$E_{\textrm{temporal}}$} $= \sum_{i=2, j, k}^{F, C, |L|} \| \myvec{x}_{i,j,k} - \myvec{x}_{i-1,j,k} \| ^2$ reduces jitter by encouraging face mesh vertices $\myvec{x}$ to remain still between neighboring frames $i-1$ and $i$.
\vspace{0.5em}

\begin{figure}[t]
\centering
\includegraphics[width=0.9\linewidth]{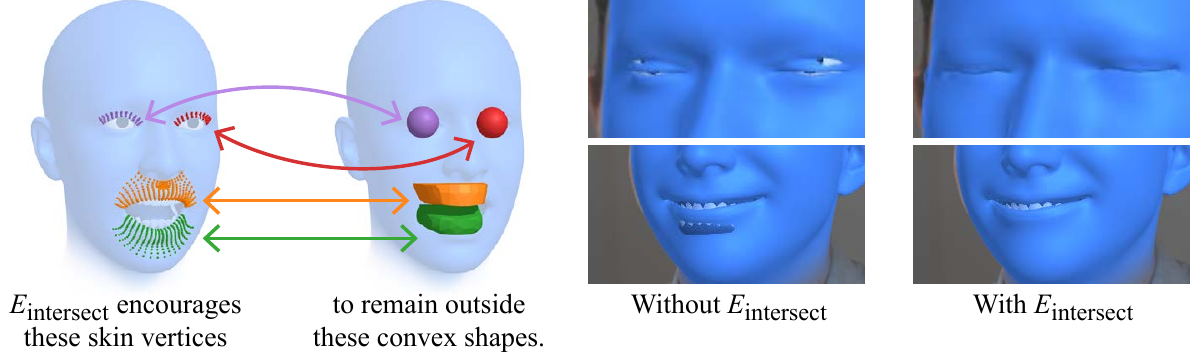}
\caption{
We encourage the optimizer to avoid face mesh self-intersections by penalizing skin vertices that enter the convex hulls of the eyeballs or teeth parts.}
\label{fig:intersection}
\end{figure}

\textbf{$E_{\textrm{intersect}}$} encourages the optimizer to find solutions without intersections between the skin and eyeballs or teeth (\autoref{fig:intersection}).
Please refer to the supplementary material for further details.

\subsection{Implementation}

We implemented two versions of our system: 
one for processing multi-camera recordings offline,
and one for real-time facial performance capture.

Our \textbf{offline} system produces the best quality results without constraints on compute.
We predict 703 landmarks with a ResNet~101~\cite{he2016deep}.
To  extract a facial Region-of-Interest (ROI) from an image we run a full-head probabilistic landmark CNN on multi-scale sliding windows, and select the window with the lowest uncertainty.
When fitting our 3DMM, we use PyTorch~\cite{NEURIPS2019_9015} to minimize $E(\myvec{\Phi})$ with L-BFGS~\cite{liu1989limited}, optimizing all parameters across all frames simultaneously.

For our \textbf{real-time} system,
we trained a lightweight dense landmark model with MobileNet V2 architecture \cite{sandler2018mobilenetv2}.
To compensate for a reduction in network capacity, we predict 320 landmarks rather than 703, and
modify the ROI strategy: aligning the face so it appears upright with the eyes a fixed distance apart.
This makes the CNN's job easier for frontal faces at the expense of profile ones.

\subsubsection{Real-time model fitting.}

We use the Levenberg-Marquardt algorithm to optimize our model-fitting energy. 
Camera and identity parameters are only fit occasionally.
For the majority of frames we fit pose and expression parameters only.
We rewrite the energy $E$ in terms of the vector of residuals, $\myvec{r}$, as 
$E(\myvec{\Phi}) = \|\myvec{r(\myvec{\Phi})}\|^2 = \sum_i r_i(\myvec{\Phi})^2$.
Then at each iteration $k$ of our optimization, we can compute $\myvec{r}(\myvec{\Phi}_k)$ and
the Jacobian, $J(\myvec{\Phi}_k) = \frac{\partial \myvec{r}(\myvec{\Phi})}{\partial \myvec{\Phi}} |^{\myvec{\Phi}=\myvec{\Phi}_k}$,
and use these to solve the symmetric, positive-semi-definite linear system, $(J^T J + \lambda \text{diag}(J^T J))\myvec{\delta}_k = -J^T\myvec{r}$ via Cholesky decomposition. 
We then apply the update rule, $\myvec{\Phi}_{k+1} = \myvec{\Phi}_k + \myvec{\delta}_k$.

In practice we do not actually form the residual vector $\myvec{r}$ nor the Jacobian matrix $J$.
Instead, for performance reasons, we directly compute the quantities $J^T J$ and $J^T \myvec{r}$
as we visit each term $r_i(\myvec{\Phi}_k)$ of the energy. 
Most of the computational cost is incurred in evaluating these products for the landmark data term, as expected.
However,
the Jacobian of landmark term residuals is not fully dense.
Each individual landmark depends on its own subset of expression parameters, and is invariant to other expression parameters.
We performed a static analysis of the sparsity of each landmark term with respect to parameters, $\partial r_i / \partial \Phi_j$, and we use this set of $i, j$ indices to reduce the cost of our outer products from $O(|\myvec{\Phi}|^2)$ to $O(m_i^2)$, where $m_i$ is the sparsified dimensionality of $\partial r_i / \partial \myvec{\Phi}$.
We further enhance the sparsity by ignoring any components of the Jacobian with an absolute value below a certain empirically-determined threshold.

By exploiting sparsity in this way, the landmark term residuals and their derivatives become very cheap to evaluate.
This formulation avoids the correspondence problem usually seen with depth images \cite{taylor2016efficient}, which requires a more expensive %
optimization. %
In addition, adding more landmarks does not significantly increase the cost of optimization.
It therefore becomes possible to implement a very detailed and well-regularized fitter with a relatively small compute burden, simply by adding a sufficient number of landmarks.
The cost of the Cholesky solve for the update $\myvec{\delta}_k$ is independent of the number of landmarks.

\section{Evaluation}

\subsection{Landmark accuracy}

We measure the accuracy of a ResNet 101 dense landmark model on the \textbf{300W}~\cite{sagonas2016threew} dataset.
For benchmark purposes only,
we employ label translation~\cite{wood2021fake} to deal with systematic inconsistencies between our 703 predicted dense landmarks
and the 68 sparse landmarks labelled as ground truth (see \autoref{tab:300W}).
While previous work~\cite{wood2021fake} used label translation to evaluate a synthetically-trained sparse landmark predictor,
we use it to evaluate a dense landmark predictor.

We use the standard normalized mean error (NME) and failure rate (FR$_{10\%}$) error metrics~\cite{sagonas2016threew}.
Our model's results in \autoref{tab:300W} are competitive with the state of the art, despite being trained with synthetic data alone.
Note: these results provide a conservative estimate of our method's accuracy as the translation network may introduce error, especially for rarely seen expressions.

\newcommand{\tikzcircle}[2][red,fill=red]{\tikz[baseline=-0.7ex]\draw[#1,radius=2pt] (0,0) circle ;}%
\newcommand{\goldmedal}{\rlap{\hspace{-7pt}\tikzcircle[gold,fill=gold]{}}}
\newcommand{\silvermedal}{\rlap{\hspace{-8pt}\tikzcircle[silver,fill=silver]{}}}
\newcommand{\bronzemedal}{\rlap{\hspace{-8pt}\tikzcircle[bronze,fill=bronze]{}}}

\begin{figure}[t]
    \begin{minipage}{6.52cm}
        \small
        \begin{tabularx}{\textwidth}{Xccc}
            & Common & Challenging & Private \\ Method & NME & NME & FR$_{10\%}$ \\
            \hline
            LAB~\cite{wayne2018lab} & \bronzemedal{}2.98 & 5.19 & 0.83 \\
            AWING~\cite{wang2019awing} & \goldmedal{}{\color{black}\textbf{2.72}} & \goldmedal{}\textbf{4.52} & \silvermedal{}0.33\\
            ODN~\cite{zhu2019odn}  & 3.56 & 6.67 & - \\
            3FabRec~\cite{browatzki20203fabrec} & 3.36 & 5.74 & \goldmedal{}\textbf{0.17} \\
            \scriptsize{Wood \etal~\cite{wood2021fake}} & 3.09 & 4.86 & \bronzemedal{}0.50 \\
            LUVLi~\cite{kumar2020luvli} &\silvermedal{}2.76 & 5.16 & -\\
            \hline
            ours {\scriptsize (L2)}  & 3.30 & \bronzemedal{}5.12 & \silvermedal{}0.33  \\
            \hline
            ours {\scriptsize (GNLL)} & 3.03 & \silvermedal{}4.80 & \goldmedal{}\textbf{0.17}  \\
        \end{tabularx}
    \end{minipage}%
    \hfill%
    \begin{minipage}{5.45cm}
    \includegraphics[width=\textwidth]{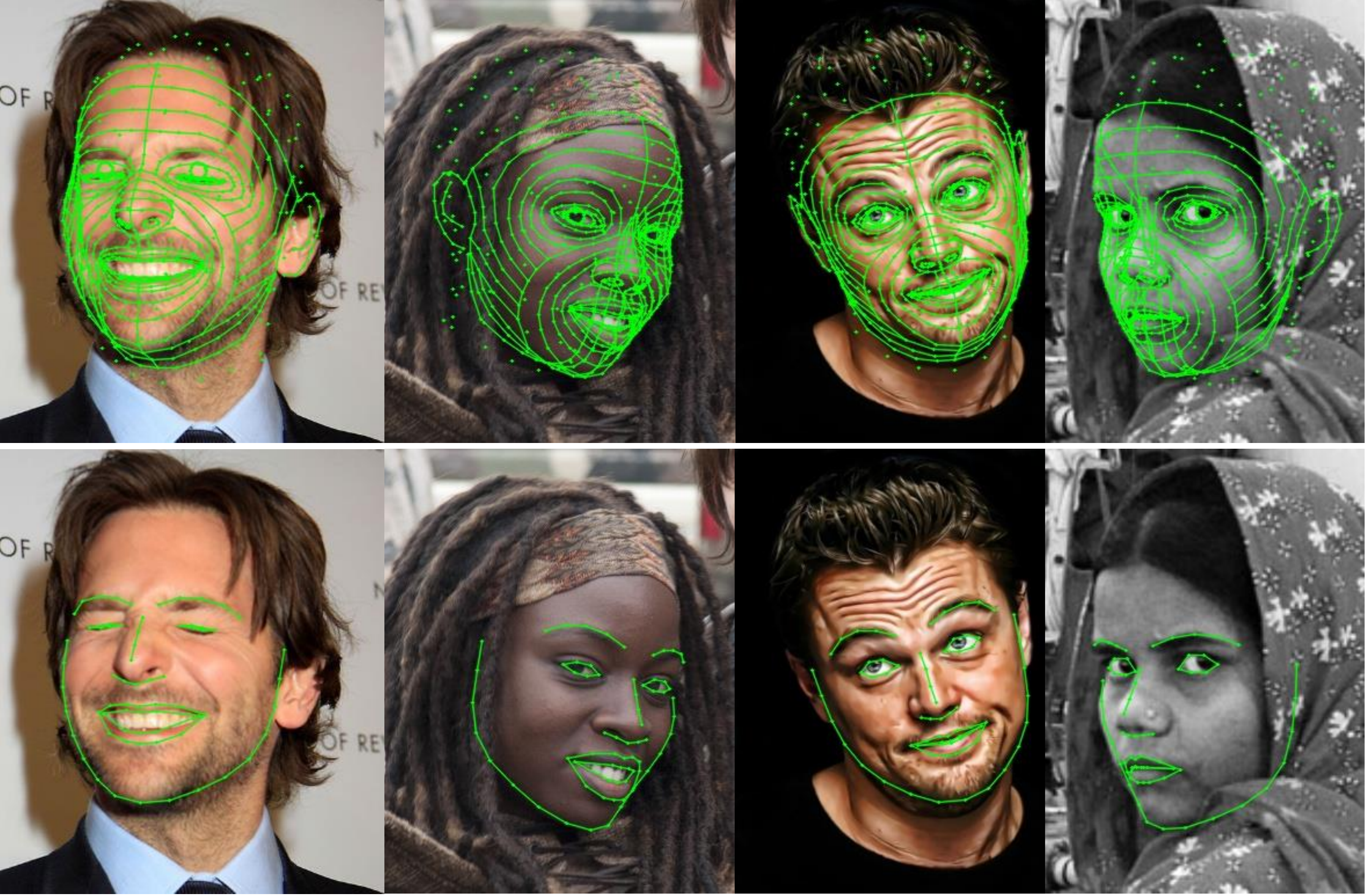}
    \end{minipage}
    \caption{Left: results on 300W dataset, lower is better. Note competitive performance of our model (despite being evaluated across-dataset) and importance of GNLL loss. Right: sample predictions (top row) with label-translated results (bottom row).}
    \label{tab:300W}
\end{figure}

\textbf{Ablation study}
We measured the importance of predicting each landmark as a random variable rather than as a 2D coordinate.
We trained two landmark prediction models,
one with our proposed GNLL loss (\autoref{eqn:loss}),
and one with a simpler L2 loss on landmark coordinate only.
Results in \autoref{tab:300W} confirm that including uncertainty in landmark regression results in better accuracy.

\begin{figure}[t]
    \hspace{2.016cm}%
    \makebox[2.016cm]{\centering MediaPipe \cite{grishchenko2020attention}}%
    \makebox[2.016cm]{\centering ours}%
    \hspace{2.016cm}%
    \hfill%
    \makebox[2.016cm]{\centering MediaPipe \cite{grishchenko2020attention}}%
    \makebox[2.016cm]{\centering ours}\\
    \includegraphics[width=\linewidth]{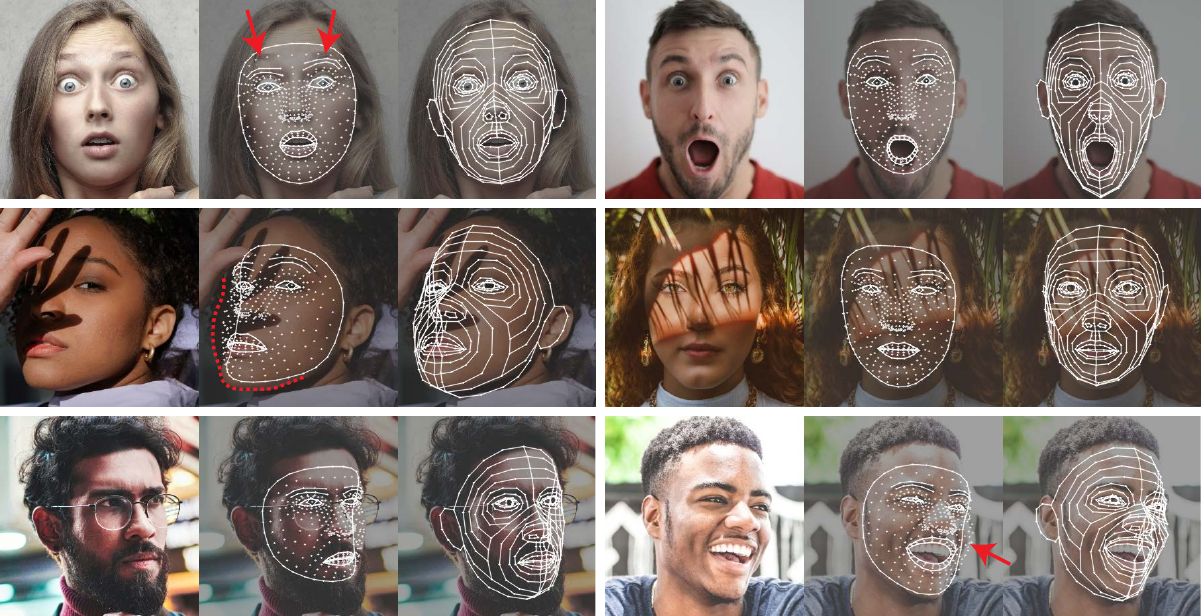}
    \caption{We compare our real-time landmark CNN (MobileNet V2) with MediaPipe Attention Mesh~\cite{grishchenko2020attention}, a publicly available method for dense landmark prediction. Our approach is more robust to challenging expressions and illumination.}
    \label{fig:experiments-mediapipe}
\end{figure}

\textbf{Qualitative comparisons}
are shown in \autoref{fig:experiments-mediapipe} between our real-time dense landmark model (MobileNet V2) and MediaPipe Attention Mesh~\cite{grishchenko2020attention}, a publicly available dense landmark method designed for mobile devices.
Our method is more robust, perhaps due to the consistency and diversity of our synthetic training data.
See the supplementary material for additional qualitative results, including landmark predictions on the Challenging subset of 300W.

\subsection{3D face reconstruction}
\label{subsec:3d-face-recon}

\begin{figure}[t]
    \begin{minipage}{6.95cm}
    \includegraphics[width=\textwidth]{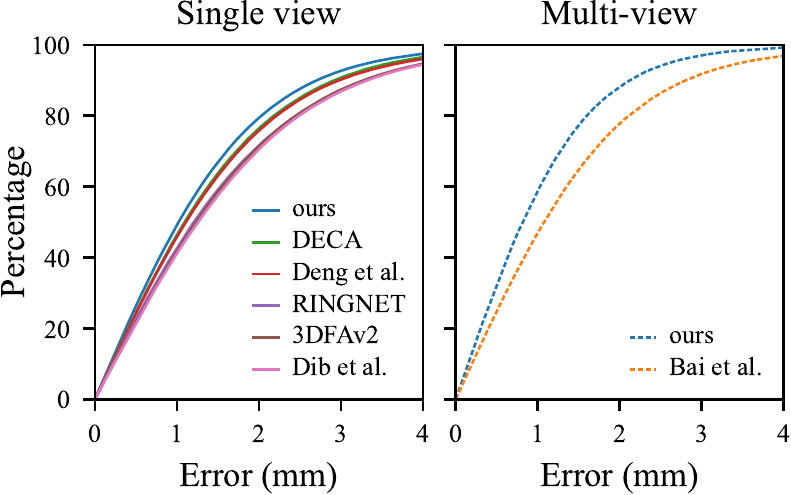}%
    \end{minipage}%
    \hfill%
    \begin{minipage}{4.95cm}
        \small
        \begin{tabularx}{\textwidth}{Xccc}
            Method & \multicolumn{3}{c}{Error (mm)}\\
            Single view & Median & Mean & Std\\
            \hline
            \scriptsize{Deng et al. \cite{deng2019accurate}} & 1.11 & 1.41 & 1.21\\ 
            RingNet \cite{RingNet:CVPR:2019} & 1.21 & 1.53 & 1.31\\ 
            3DFAv2 \cite{guo2020towards} & 1.23 & 1.57& 1.39\\
            DECA \cite{Feng:SIGGRAPH:2021} & 1.09 & 1.38 & 1.18\\
            Dib et al. \cite{dib2021towards} & 1.26 & 1.57 & 1.31\\ 
            \textbf{ours} & \textbf{1.02} & \textbf{1.28} & \textbf{1.08}\\[0.5em]
            Multi-view &  &  & \\
            \hline
            Bai et al. \cite{Bai_2021_CVPR} & 1.08 & 1.35 & 1.15\\ 
            \textbf{ours} & \textbf{0.81} & \textbf{1.01} & \textbf{0.84}\\
        \end{tabularx}
    \end{minipage}
    \caption{Results for the NoW Challenge \cite{RingNet:CVPR:2019}. We outperform the state of the art on both single- and multi-view 3D face reconstruction.}
    \label{fig:experiments-now-results}
\end{figure}

Quantitatively, we compare our offline approach with recent methods on two benchmarks: the NoW Challenge~\cite{RingNet:CVPR:2019} and the MICC dataset~\cite{Bagdanov:2011:FHF:2072572.2072597}. %

\textbf{The NoW Challenge}~\cite{RingNet:CVPR:2019} provides a standard evaluation protocol for measuring the accuracy and robustness of 3D face reconstruction in the wild.
It consists of 2054 face images of 100 subjects along with a 3D head scan for each subject which serves as ground truth. %
We undertake the challenge in two ways:
\textit{single view}, where we fit our face model to each image separately, and 
\textit{multi-view}, where we fit a per-subject face model to all image of a particular subject.
As shown in \autoref{fig:experiments-now-results}, we achieve state of the art results. %

\textbf{The MICC dataset}~\cite{Bagdanov:2011:FHF:2072572.2072597} consists of 3D face scans and videos of 53 subjects.
The videos were recorded in three environments: a ``cooperative'' laboratory environment, an indoor environment, and an outdoor environment.
We follow Deng et al.~\cite{deng2019accurate}, and evaluate our method in two ways:
\textit{single view}, where we estimate one face shape per frame in a video, and average the resulting face meshes,
and
\textit{multi-view}, where we fit a single face model to all frames in a video jointly.
As shown in \autoref{tab:florence}, we achieve state of the art results.

Note that many previous methods are incapable of aggregating face shape information across multiple views.
The fact ours can benefit from multiple views highlights the flexibility of our hybrid model-fitting approach.

\begin{table}[t]
    \caption{Results on the MICC dataset~\cite{Bagdanov:2011:FHF:2072572.2072597}, following the single and multi-frame evaluation protocol of Deng \etal \cite{deng2019accurate}. We achieve state-of-the-art results.}
    \label{tab:florence}
    \small
    \begin{tabularx}{0.45\textwidth}[t]{Xccc} %
        Method & \multicolumn{3}{c}{Error (mm), mean}\\
        Single view & Coop. & Indoor & Outdoor\\
        \hline
        Tran \etal \cite{tuan2017regressing} & 1.97 & 2.03 & 1.93\\ 
        \scriptsize{Genova \etal \cite{genova2018unsupervised}} & 1.78 & 1.78 & 1.76\\ 
        Deng \etal \cite{deng2019accurate} & 1.66 & 1.66 & 1.69\\
        \textbf{ours} & \textbf{1.64} & \textbf{1.62} & \textbf{1.61}
    \end{tabularx}%
    \hfill%
    \begin{tabularx}{0.54\textwidth}[t]{Xccc} %
        Method & \multicolumn{3}{c}{Error (mm), mean}\\
        Multi-view & Coop. & Indoor & Outdoor\\
        \hline
        \scriptsize{Piotraschke and Blanz \cite{piotraschke2016automated}} & 1.68 & 1.67 & 1.72\\ 
        Deng \etal \cite{deng2019accurate} & 1.60 & 1.61 & 1.63\\ 
        \textbf{ours} & \textbf{1.43} & \textbf{1.42} & \textbf{1.42}
    \end{tabularx}%
\end{table}

\subsubsection{Ablation studies} 
\begin{figure}[t]
    \begin{minipage}{6.95cm}
    \makebox[0.25\textwidth]{Fit with:}%
    \makebox[0.25\textwidth]{\centering 68 ldmks.}%
    \makebox[0.25\textwidth]{\centering 320 ldmks.}%
    \makebox[0.25\textwidth]{\centering 703 ldmks.}\\[0.1em]
    \includegraphics[width=0.25\textwidth]{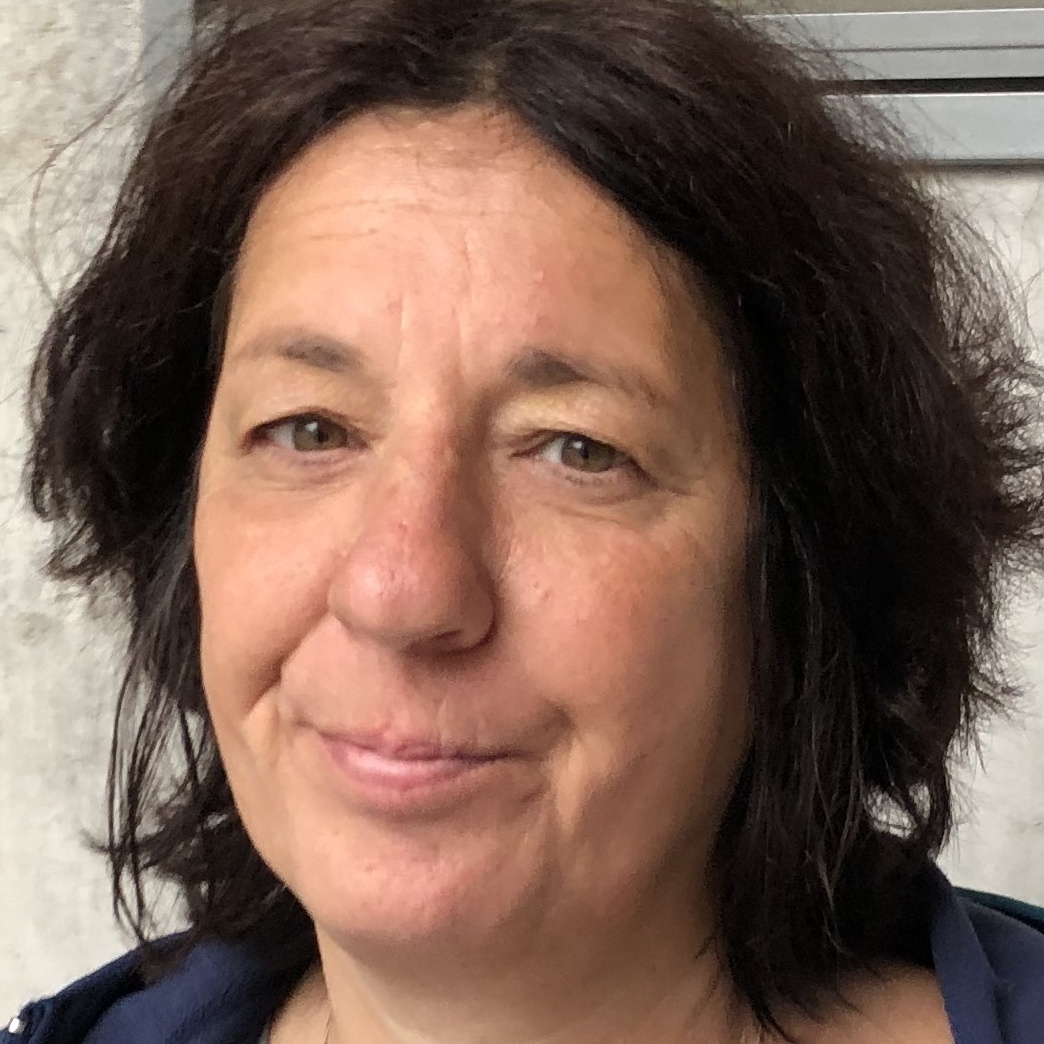}%
    \includegraphics[width=0.25\textwidth]{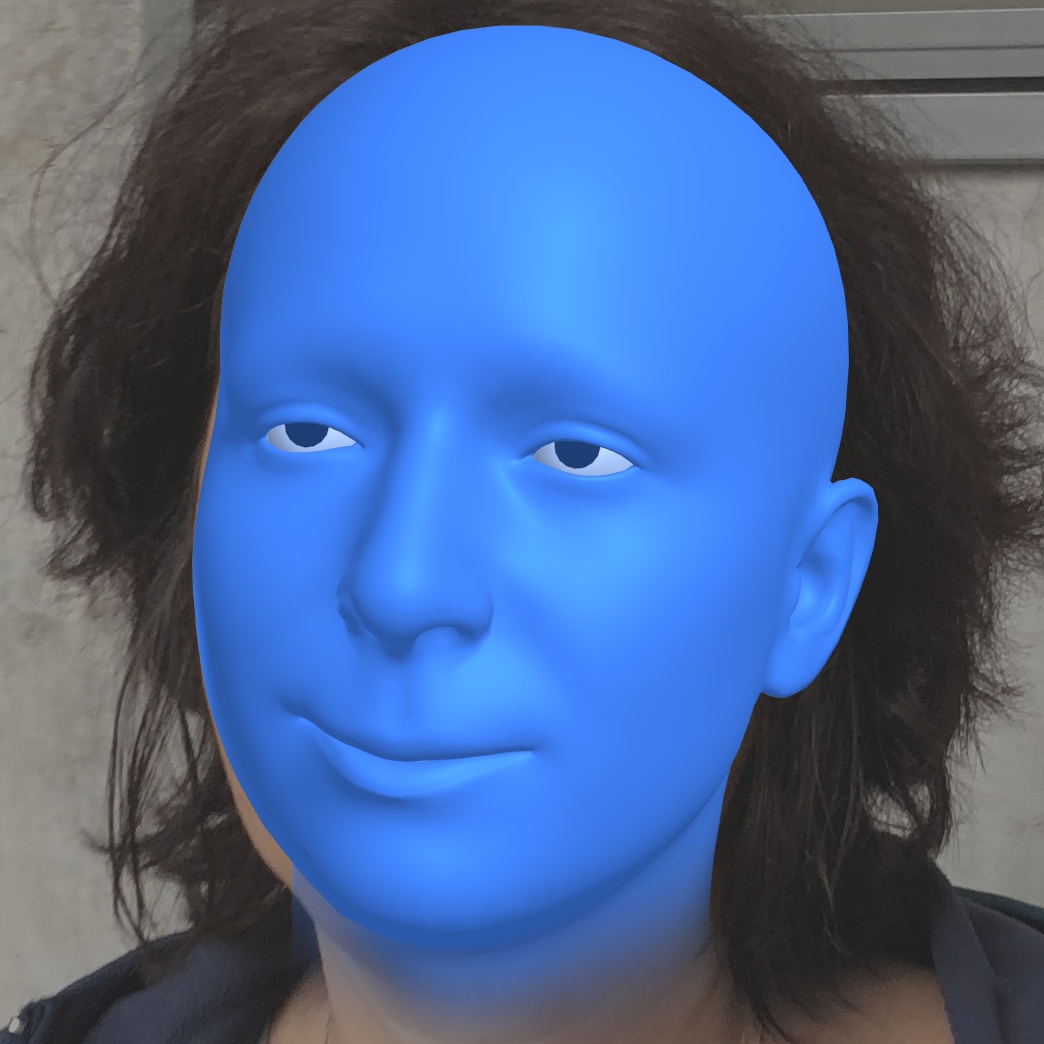}%
    \includegraphics[width=0.25\textwidth]{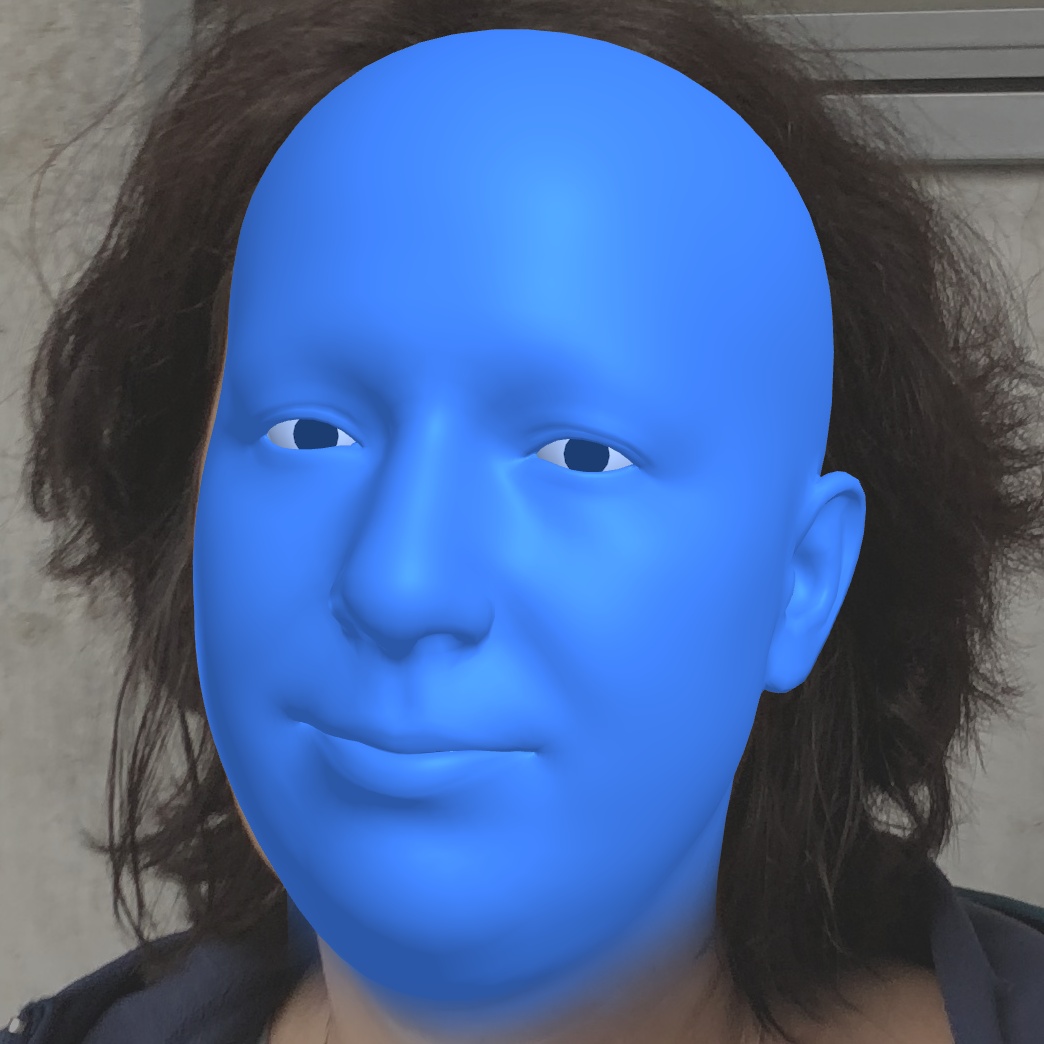}%
    \includegraphics[width=0.25\textwidth]{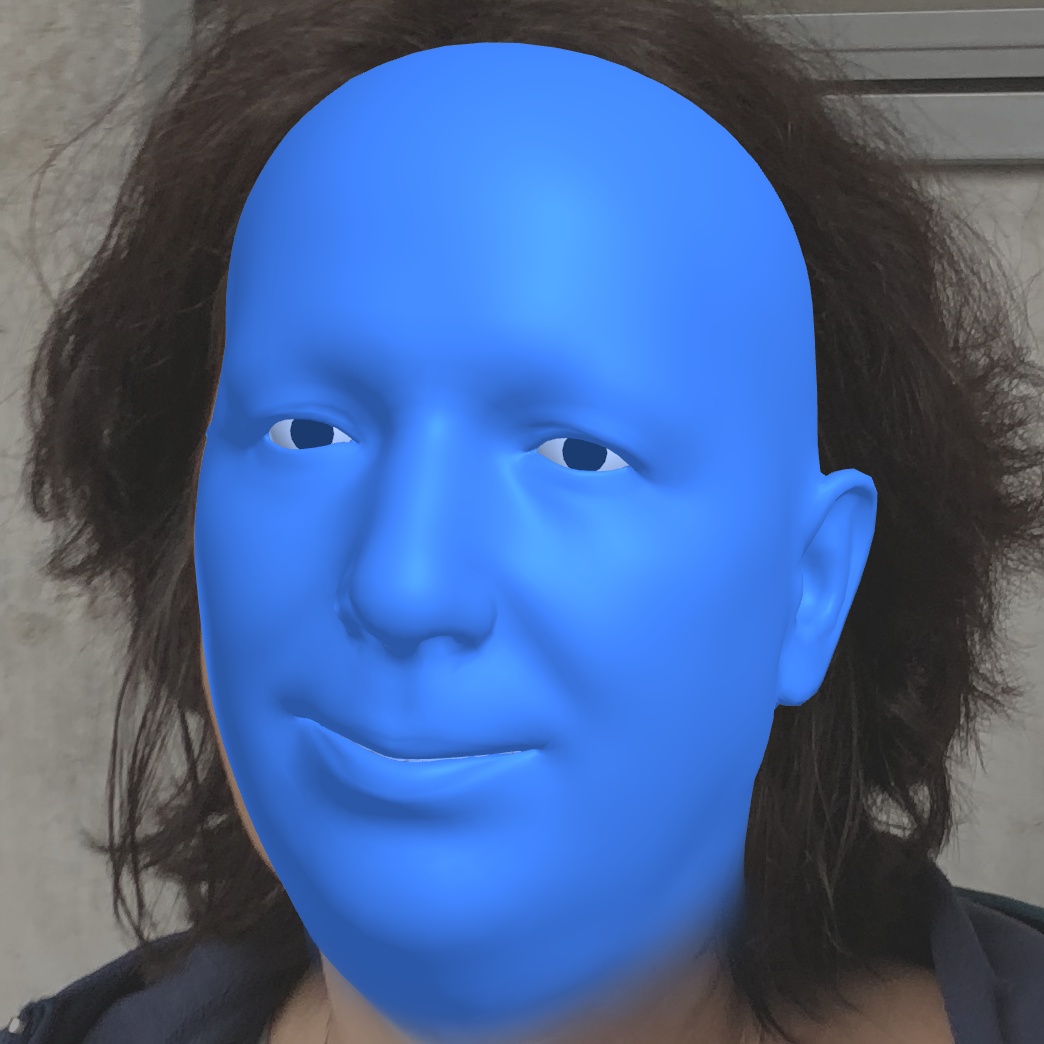}
    \end{minipage}%
    \hfill%
    \begin{minipage}{4.95cm}
        \small
        \begin{tabularx}{\textwidth}{Xccc}
            Number of & \multicolumn{3}{c}{Error (mm)}\\
            Landmarks & Median & Mean & Std\\
            \hline
            68 & 1.10 & 1.38 & 1.16\\ 
            320 & 1.00 & 1.24 & 1.02\\ 
            \textbf{703} & \textbf{0.95} & \textbf{1.17} & \textbf{0.97}\\%[0.3em]
            \hline
            703 {\scriptsize(without $\sigma$)} & 1.02 & 1.26 & 1.03\\
        \end{tabularx}
    \end{minipage}
    \caption{Ablation studies on the NoW \cite{RingNet:CVPR:2019} validation set confirm that denser is better: model fitting with more landmarks leads to more accurate results.
    In addition, we see that fitting without using $\sigma$ leads to worse results.}
    \label{fig:experiments-now-ablation}
\end{figure}

We conducted an experiment to measure the importance of landmark quantity for 3D face reconstruction.
We trained three landmark CNNs, predicting 703, 320, and 68 landmarks respectively, and 
used these on the NoW Challenge (validation set).
As shown in \autoref{fig:experiments-now-ablation},
fitting with more landmarks results in more accurate 3D face reconstruction.

In addition, we investigated the importance of using landmark uncertainty $\sigma$ in model fitting.
We fit our model to 703 landmark predictions on the NoW validation set,
but using fixed rather than predicted $\sigma$.
\autoref{fig:experiments-now-ablation} (bottom row of table) shows that
fitting without $\sigma$ leads to worse results.

\subsubsection{Qualitative comparisons} 
between our work and several publicly available methods
\cite{RingNet:CVPR:2019,deng2019accurate,guo2020towards,shang2020self,Feng:SIGGRAPH:2021}
can be found in \autoref{fig:experiments-3d-recon-qaul}.

\subsection{Facial performance capture}

\subsubsection{Multi-view}

\begin{figure}[t]
    \includegraphics[width=\textwidth]{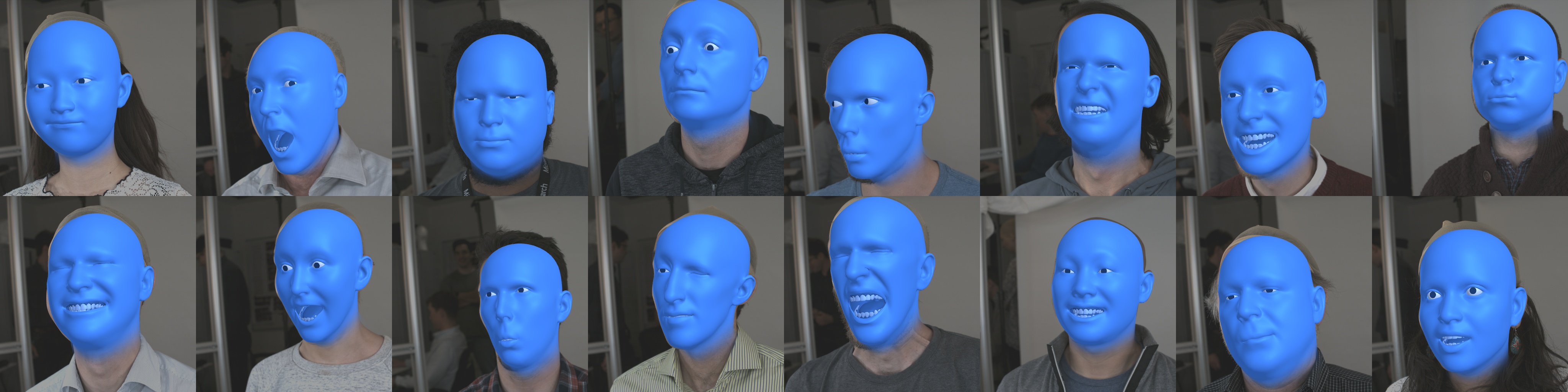}%
\caption{We demonstrate the robustness and reliability of our method by using it to collect a massive database of 125,000 facial expressions, fully automatically.}
\label{fig:eval-multi-view-perf-cap}
\end{figure}

Good synthetic training data requires a database of facial expression parameters from which to sample.
We acquired such a database by conducting markerless facial performance capture for 108 subjects.
We recorded each subject in our 17-camera studio, and processed each recording with our offline multi-view model fitter.
For a 520 frame sequence it takes 3 minutes to predict dense landmarks for all images, and a further 9 minutes to optimize face model parameters.
See \autoref{fig:eval-multi-view-perf-cap} for some of the 125,000 frames of expression data captured with our system.
As the system which is used to create the database is then subsequently re-trained with it, we
produced several databases in this manner until no further improvement was seen.
We do not reconstruct faces in fine detail like previous multi-view stereo approaches \cite{globallyConsistentReconstructionPopa, passiveFacialPerfCapture, anchorFramesPaper}.
However, while previous work can track a detailed 3D mesh over a performance, our approach reconstructs the performance with richer semantics: identity and expression parameters for our generative model.
In many cases it is sufficient to reconstruct the low-frequency shape of the face accurately, without fine details.

\subsubsection{Real-time monocular}

\begin{figure}[t]
    \scriptsize
    \begin{tabularx}{\textwidth}{YYYYYYYY}
        & RingNet
        & Deng et al.
        & 3DFAv2
        & MGCNet 
        & DECA 
        & ours
        & ours\\
        & \cite{RingNet:CVPR:2019}
        & \cite{deng2019accurate}
        & \cite{guo2020towards}
        & \cite{shang2020self}
        & \cite{Feng:SIGGRAPH:2021}
        & (offline)
        & (real-time)
    \end{tabularx}
    \includegraphics[width=\linewidth]{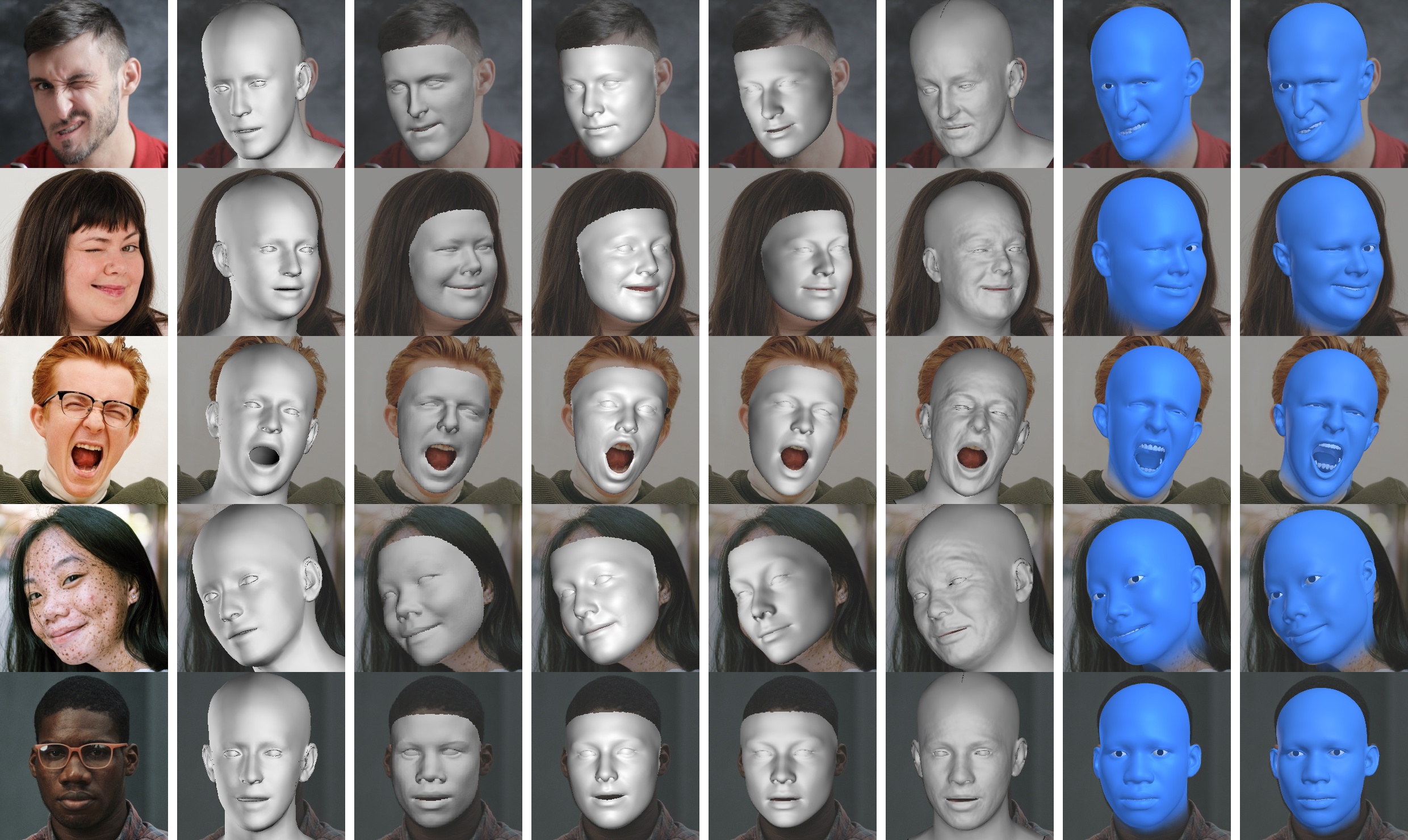}
    \caption{
Compared to previous recent monocular 3D face reconstruction methods,
ours better captures gaze, expressions like winks and sneers, and subtleties of facial identity.
In addition, our method can run in real time with only a minor loss of fidelity.}
    \label{fig:experiments-3d-recon-qaul}
\end{figure}

See the last two columns of \autoref{fig:experiments-3d-recon-qaul} for a comparison between our offline and real-time systems for monocular 3D model-fitting.
While our offline system produces the best possible results by using a large CNN and optimizing over all frames simultaneously,
our real-time system can still produce accurate and expressive results fitting frame-to-frame.
Please refer to the supplementary material for more results. %
Running on a single CPU thread (i5-11600K), our real-time system spends 6.5ms processing a frame (150FPS), of which 4.1ms is spent predicting dense landmarks and 2.3ms is spent fitting our face model.

\begin{figure}[t]
    \includegraphics[width=\textwidth]{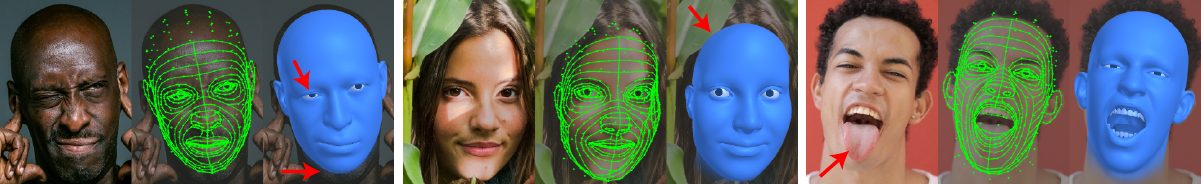}%
\caption{Bad landmarks result in bad fits, and we are incapable of tracking the tongue.}
\label{fig:limitations}
\end{figure}

\section{Limitations and future work}

Our method depends entirely on accurate landmarks.
As shown in \autoref{fig:limitations}, if landmarks are poorly predicted, the resulting model fit suffers.
We plan to address this by improving our synthetic training data.
Additionally, since our model does not include tongue articulation we cannot recover tongue movement.

Heatmaps have dominated landmark prediction for some time~\cite{bulat2017far,bulat2021subpixel}.
We were pleasantly surprised to find that directly regressing 2D landmark coordinates with unspecialized architectures works well and eliminates the need for computationally-costly heatmap generation.
In addition, we were surprised that predicting $\sigma$ helps accuracy. %
We look forward to further investigating direct probabilistic landmark regression as an alternative to heatmaps in future work.

In conclusion,
we have demonstrated that dense landmarks are an ideal signal for 3D face reconstruction.
Quantitative and qualitative evaluations have shown that our approach outperforms those previous by a significant margin,
and excels at multi-view and monocular facial performance capture. %
Finally, our approach is highly efficient, and runs at over 150FPS on a single CPU thread.

\subsubsection*{Acknowledgements}
Thanks to Jiaolong Yang for sharing the code from Deng et al.~\cite{deng2019accurate}, and Timo Bolkart for help with the NoW Challenge evaluation \cite{RingNet:CVPR:2019}.

\clearpage

\section*{Supplementary material}

\subsection*{Additional qualitative results}

\begin{figure}[t]
\includegraphics[width=\textwidth]{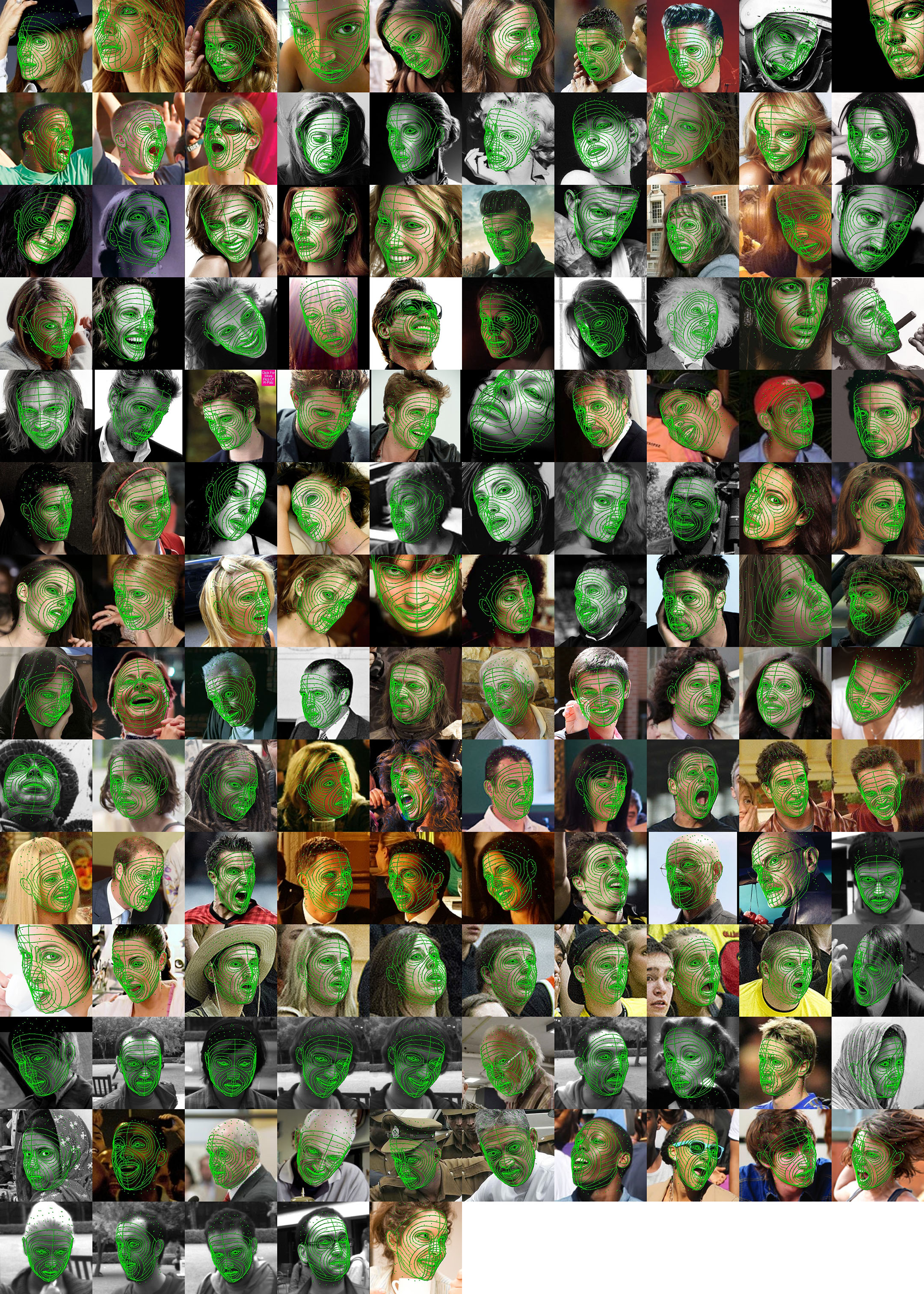}
\caption{Dense landmark predictions for all of the images in 300W Challenging dataset. Note the accuracy of our landmark model in challenging scenarios including extreme expressions, occlusion, pose variation, lighting variation, and poor image quality.}
\label{fig:suppl-300w}
\end{figure}

Please see \autoref{fig:suppl-300w} for dense landmark predictions on the entire 300W Challenging dataset, using a ResNet 101 \cite{he2016deep}.

\begin{figure}[t]
    \hspace{0.1428\textwidth}%
    \makebox[0.1428\textwidth]{\centering \scriptsize{RingNet \cite{RingNet:CVPR:2019}}}%
    \makebox[0.1428\textwidth]{\centering \scriptsize{Deng et al. \cite{deng2019accurate}}}%
    \makebox[0.1428\textwidth]{\centering \scriptsize{3DFAv2 \cite{guo2020towards}}}%
    \makebox[0.1428\textwidth]{\centering \scriptsize{MGCNet \cite{shang2020self}}}%
    \makebox[0.1428\textwidth]{\centering \scriptsize{DECA \cite{Feng:SIGGRAPH:2021}}}%
    \makebox[0.1428\textwidth]{\centering \scriptsize{Ours}}\\[0.07em]
    \includegraphics[width=\textwidth]{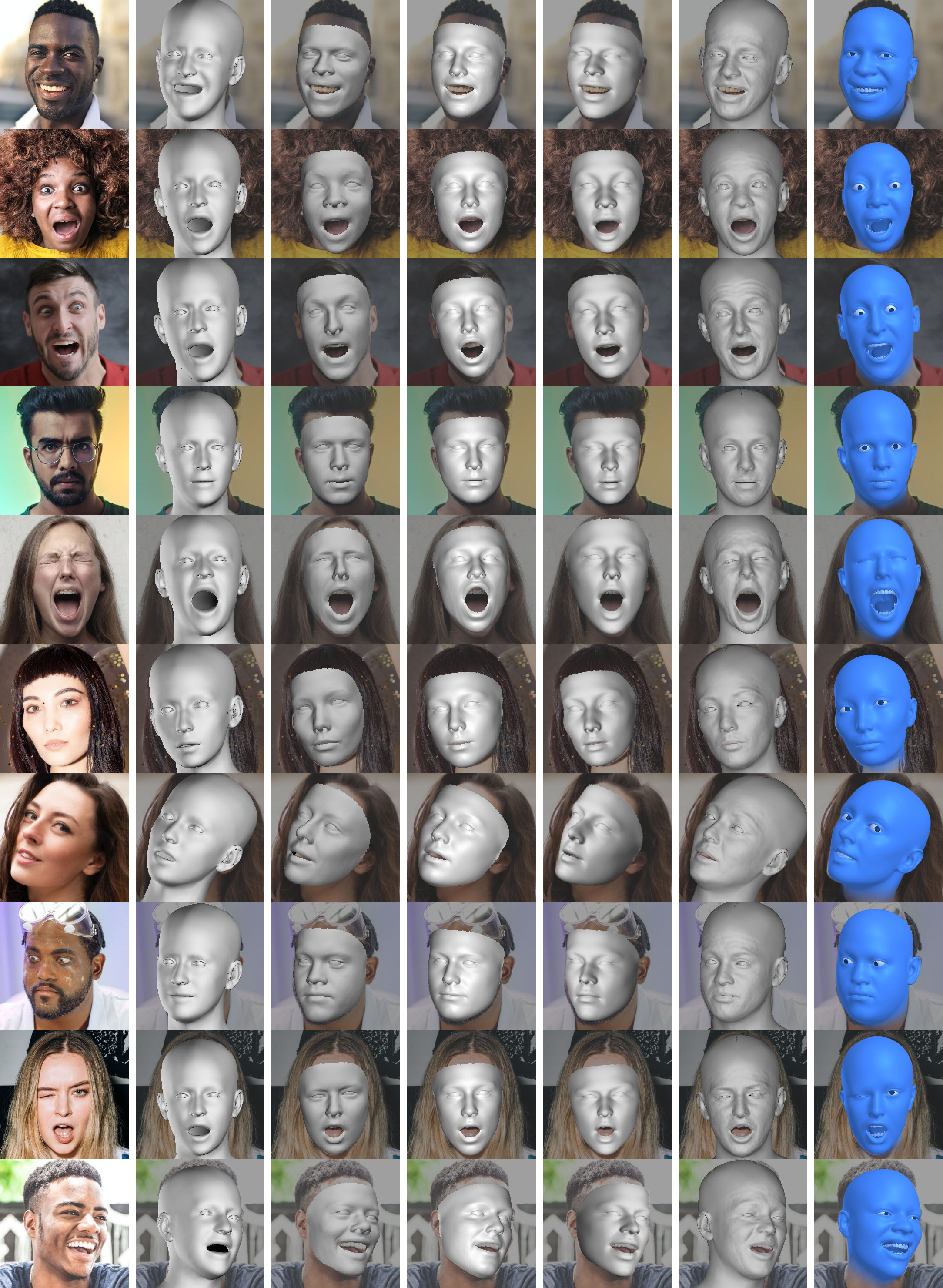}
    \caption{Further qualitative comparisons between our approach and publicly available recent previous methods for monocular 3D face reconstruction.}
    \label{fig:suppl-3d-face-recon-qual}
\end{figure}

Please see \autoref{fig:suppl-3d-face-recon-qual} for qualitative comparisons between our approach and several recent methods for 3D face reconstruction in the wild.

\subsection*{$E_{\textrm{intersect}}$}

Since our 3D face model contains separate parts for the teeth and eyeballs, intersections can occur.
Though they are uncommon,
we encourage the optimizer to avoid face mesh self-intersections by penalizing skin vertices that enter the convex hulls of the eyeballs or teeth parts.
$$E_{\textrm{intersect}} = E_{\textrm{eyeballs}} + E_{\textrm{teeth}}$$
We attach a sphere of fixed radius to each eyeball center.
For each eyelid skin vertex that falls inside its corresponding eyeball sphere, $E_{\textrm{eyeballs}}$ penalizes the squared distance between that vertex to the sphere's exterior surface.
Since this is trivial to implement, we will omit the details.

However, the teeth cannot be well-represented with a simple primitive like a sphere, so $E_{\textrm{teeth}}$ is more complicated.
Instead, we represent the upper and lower teeth parts each with a convex hull of $J$ planes defined by normal vector $\hat{\myvec{n}}_j$ and distance to origin $p_j$.
Lets say $I$ represents a set of lip vertices we wish to keep outside one of these convex hulls.
$$
E_{\textrm{teeth}} = \sum_{i \in I} D_i^2
$$
Where $D_i$ measures the distance the $i$\textsuperscript{th} skin vertex is inside the convex hull,
$$
D_{i} = \min_{j \in J} \left\{ d_{i,1}, \ldots, d_{i,J} \right\}
$$
and $d_{i,j}$ measures the internal distance between the $i$\textsuperscript{th} skin vertex and the $j$th plane of the convex hull.
$$
    d_{i,j} = -\textrm{min} \left( \hat{\myvec{n}}_j \cdot \myvec{x}_i + p_j, 0 \right)
$$

\subsection*{Region-of-Interest extraction}

\begin{figure}[t]
    \includegraphics[width=\linewidth]{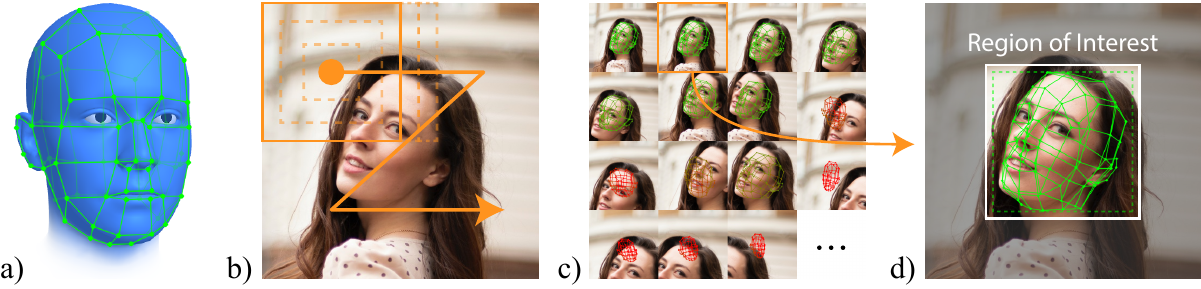}
    \caption{
    To extract a distortion-free Region-of-Interest (ROI) of the head we run a full-head sparse landmark CNN (a) on multi-scale sliding windows across the full image (b), take the landmarks from the most confident window (c), inscribe an expanded square around them, and use it as our ROI (d) for our dense landmark CNN.
    }
    \label{fig:method-roi-extraction-1}
\end{figure}

\begin{figure}[t]
    \hfill
    a) \includegraphics[width=0.25\linewidth]{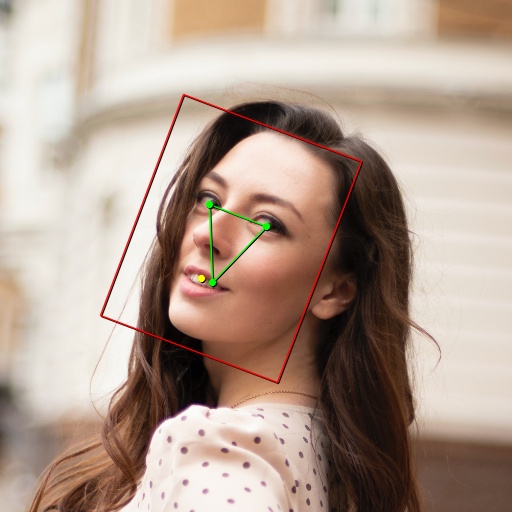} \hspace*{1.25cm}
    b) \includegraphics[width=0.25\linewidth]{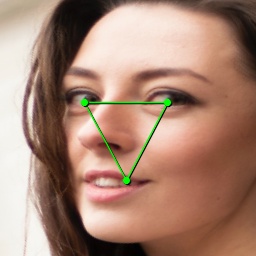}
    \hspace*{\fill}%
    \caption{
    When running in real time, we extract ROIs using an affine transform. The eyes and mouth, shown as green points in (a) are remapped every frame to a fixed triangle in ROI space (b). The resulting rotated ROI rectangle, shown in red in (a) is remapped into the square (b).
    }
    \label{fig:method-roi-extraction-2}
\end{figure}

Our method is ``top-down'', so it relies on being provided a reasonable Region-of-Interest (ROI) of a face.
During offline processing, where compute is not paramount, we prefer to extract distortion-free ROIs from a full image, with zero rotation and uniform scaling (see \autoref{fig:method-roi-extraction-1}).

When running in real time, we modify our ROI strategy to perform more data normalization, producing ROIs where the eyes always appear in the same place, and the mouth always appears on the same horizontal line (see \autoref{fig:method-roi-extraction-2}).
This corresponds to a rotation and non-uniform scaling.
While this makes the task easier for the dense landmark CNN (preferable for low-capacity neural networks), it breaks down for profile faces, so we prefer the distortion free approach described in \autoref{fig:method-roi-extraction-1} when compute is not constrained.

\clearpage

\subsection*{Additional implementation details}
\subsubsection*{Training details.}
Both (a) the 703 landmarks model with pretrained ResNet 101 \cite{he2016deep} backbone for offline multi-view fitting and (b) the 320 landmarks model with pretrained MobileNet V2 \cite{sandler2018mobilenetv2} backbone for real-time monocular fitting, were trained with batch size 128 and learning rate schedule StepLR(step\_size=100, gamma=0.5).
We used AdamW \cite{ADAMW} in the PyTorch library \cite{NEURIPS2019_9015} with default optimizer hyperparameters.
We trained the models for 300 epochs and picked the model from the epoch with the lowest validation error on synthetic data.

\subsubsection*{Focal length initialization and optimization.}
If the focal length of the camera is known, we keep it fixed during 3D model fitting.
Otherwise, if it is unknown, we initialize it to be 45 degree effective horizontal FOV, and optimize it together with other 3D face model parameters.
For simplicity, we assume the principal point is at image center and focal length value (in pixels) is the same for x and y direction (square pixels)
Empirically, we found that these assumptions were acceptable.

\clearpage

\bibliographystyle{splncs04}
\bibliography{bib}
\end{document}